\documentclass[trsc,nonblindrev]{informs3} 

\OneAndAHalfSpacedXI 

\usepackage[authoryear,round]{natbib}
\usepackage[colorlinks=true,citecolor=blue]{hyperref}
\usepackage[T1]{fontenc}

 \bibpunct[, ]{(}{)}{,}{a}{}{,}%
 \def\bibfont{\small}%
\usepackage{tabularx}
\usepackage{booktabs}
\usepackage{multirow}
\usepackage{makecell}
\usepackage{svg}
\usepackage{algorithm}
\usepackage{subcaption}
\usepackage{algpseudocode}
\usepackage{algorithm, algpseudocode}
\usepackage{booktabs}
\newtheorem{theorem}{Theorem}

\newtheorem{definition}{Definition}

\EquationsNumberedThrough    

\begin{document}



\RUNTITLE{Physics-constrained IF}

\TITLE{Understanding Data Influence in Transportation Networks: A Physics-Constrained Influence-Function Framework}


\ARTICLEAUTHORS{
\AUTHOR{Hamidreza Attari}
\AFF{Department of Civil and Environmental Engineering, University of California, Davis, CA, USA,  \EMAIL{hattari@ucdavis.edu}}
\AUTHOR{Yueyue Fan}
\AFF{Department of Civil and Environmental Engineering, University of California, Davis, CA, USA, \EMAIL{yyfan@ucdavis.edu} \\
Corresponding author}
\AUTHOR{Johannes O. Royset}
\AFF{Daniel J. Epstein Dept. of Industrial and Systems Engineering, University of Southern California, CA, USA, \EMAIL{royset@usc.edu}}}


\maketitle

\textbf{Abstract:} Understanding how individual training observations influence model predictions is important for developing reliable and interpretable data-driven transportation systems. Existing influence-function methods, however, are primarily formulated for unconstrained learning and do not explicitly account for physical constraints that couple different components of a transportation network. Under such constraints, perturbing a training observation can affect model parameters and predictions only through physically feasible directions, potentially changing how its influence propagates across the network. This paper develops a hard physics-informed influence-function framework for constrained learning, in which training-data influence is characterized within the physically feasible space defined by network constraints. The analytical results reveal how physical constraints alter the magnitude, direction, and spatial distribution of training-sample influence relative to conventional influence measures. Using traffic prediction under flow-conservation constraints as a case study, we show that influential observations identified by the proposed framework differ systematically from those identified by classical influence functions. Through orthogonal decomposition, we quantitatively characterize how much of a perturbation in the training data is propagated through the network via the flow-conservation constraints and how much is removed by the constraint projection. The proposed influence framework provides a physically consistent approach for evaluating training-data influence in constrained transportation learning problems, with potential applications to data diagnostics, sensor evaluation, and targeted data collection.


%


\section{Introduction}

The rapid growth of AI and machine learning is creating new opportunities to develop data-driven models for understanding, predicting, and managing complex transportation systems. The accuracy and robustness of these models heavily depend on the quality and informativeness of the observations used for training. In transportation systems, data are often incomplete, noisy, and unevenly distributed across the network. Therefore, understanding how training data influence model learning and performance becomes increasingly important as such understanding can help identify where additional data collection is needed or matters most.

In the machine learning literature, Influence Functions (IFs), originating from robust statistics, have been adapted as a principled framework for quantifying the influence of training data. IFs provide a useful means of understanding model behavior by characterizing model sensitivity to perturbations in training data and identifying influential observations; such influence measures can further provide a basis for assessing the value of training information \citep{koh17a}. However, classical IF formulations primarily consider unconstrained learning, where observations influence model parameters through the empirical loss \citep{PolinaIF, basu2021influence}. This limits the direct applicability of classical IF methods to transportation network problems, where learning is often subject to physical constraints that capture the spatial and temporal interdependencies among network components. Such constraints fundamentally alter how model parameters respond to perturbations in training data, requiring influence analysis that explicitly accounts for the constrained learning process. 


In this paper, we establish a new framework for evaluating data influence in transportation networks by approaching the problem as a physics-constrained sensitivity problem.   While the proposed influence-function framework is motivated by the broader class of physics-constrained learning problems, network-level physical constraints are particularly fundamental to transportation systems. Many transportation problems are defined over networks in which observations and system states across different components are coupled through physical and operational constraints such as flow conservation,  network connectivity, and capacity restrictions. We therefore use short-term transportation network flow prediction as a motivating use case for developing the proposed framework. In this setting, flow-conservation constraints provide a concrete representation of network-level coupling among link flows, allowing us to demonstrate why such structural dependencies should be explicitly incorporated into influence-function analysis.

Physics-informed machine learning has received growing attention in both the broader AI community and transportation research, reflecting increasing recognition that incorporating physical knowledge and system structure into learning can improve model accuracy, robustness, and consistency with underlying system behavior.  For example, in the transportation community, early efforts in short-term traffic prediction progressed from classical statistical and machine-learning methods to increasingly sophisticated deep learning architectures that capture temporal and spatial dependencies in traffic data \citep{arima, MA2015187, Fu7804912}. More recent graph-based approaches, such as the diffusion convolutional recurrent neural network (DCRNN) \citep{Li2017DiffusionCR} and the spatio-temporal graph convolutional network (STGCN) \citep{ijcai2018p505}, incorporate road-network topology into the learning architecture, allowing dependencies among connected network components to be captured. However, representing network connectivity does not necessarily enforce the physical relationships governing flows over the network. Despite their strong predictive performance, these models generally do not explicitly impose network-level physical constraints, such as flow conservation, on model learning and prediction. Predictions that are physically implausible can misrepresent the actual system state, trigger ineffective or counterproductive downstream actions, and ultimately impair the performance and reliability of transportation networks. 

At the methodological level, physics-informed and constrained learning approaches have increasingly shifted from soft constraint enforcement toward hard feasibility guarantees. Early approaches typically incorporated physical or domain knowledge through penalty terms in the training loss, including penalty weights or Lagrange multipliers, thereby encouraging—but not guaranteeing—constraint satisfaction \citep{Dener2020TrainingNN, MrquezNeila2017ImposingHC}. More recent approaches seek to enforce constraints directly within neural networks through feasible-set parameterizations, differentiable projections, repair mechanisms, KKT-based formulations, or variable transformations \citep{min2025hardnethardconstrainedneuralnetworks, inproceedings, huang2021differentiableprojectionconstraineddeep, pmlr-v202-wang23at, 2023arXiv230708336, articleKonstantinov, chen2024physicsinformedneuralnetworkshard}. Collectively, these developments demonstrate growing recognition that physical constraints should be explicitly accounted for in how AI models learn and generate predictions. Yet, while physical constraints have increasingly been incorporated into how models learn, their implications for how the influence of training data is measured remain largely unexplored.

Existing sensitivity analysis and interpretability methods, including perturbation-based approaches and IFs, have primarily been developed for unconstrained learning settings, where prediction sensitivity reflects how changes in input data propagate through the learned mapping from observations to outputs. However, when hard physical constraints are embedded into the learning architecture, the prediction response is no longer determined solely by learned data-driven relationships. Instead, it is shaped by the interaction between learned representations and the feasible space defined by physical constraints. As a result, perturbations in input data may influence predictions through both the learned model and the constraint-enforcing mechanism, creating new challenges for identifying the true sources of predictive information. Understanding which sensor observations contribute most to physically constrained predictions is critical for improving data quality management and guiding sensor maintenance, calibration, and investment decisions in transportation networks.

In summary, existing literature shows that spatiotemporal deep learning can deliver strong forecasting results. Hard constraints can ensure physical consistency by design. IFs can connect predictions to individual training observations. However, these three areas have not been linked. As shown in table \ref{tab:related_work}, despite growing interest in IFs, limited attention has been given to integrating them with learning architectures that impose hard physical constraints, tracing training-data influence through constraint projection, or using the resulting influence scores to identify sensors that disproportionately affect predictions from physically constrained models. 

To address those gaps, this paper introduces a hard physics-informed influence-function framework for learning under physical constraints. The proposed framework quantifies training-data influence within the physically feasible space through constraint projection, explicitly capturing how network-level physical constraints alter the magnitude, distribution, and value of information.  We first develop a physics-informed spatiotemporal neural network for short-term traffic flow forecasting. The model combines data-driven learning with a Hard Constraint Projection layer. This layer projects raw neural network outputs onto a set defined by junction-level flow conservation. We then analyze how local perturbations in traffic observations affect the learned prediction. To do this, we create a hard physics-informed influence-function framework that improves on traditional influence analysis for constrained traffic learning. Unlike standard influence functions, which quantify training-data influence on the unconstrained model output, our framework evaluates influence on the physically feasible output obtained through constraint projection.  As a result, the influence scores show how physical constraints change, reduce, or redistribute the impact of training samples across the network. Practically, finding these influence scores can help identify critical infrastructure in transportation networks. 


In the rest of the paper, Section 2 explains our physics constrained influence analysis method, Section 3 describes the setup of numerical experiments in short-term traffic prediction context, Section 4 presents results, and Section 5 summarizes key conclusions and discusses possible future extensions.

\begin{table*}[h]
\centering
\caption{Related work organized by how physics is handled and how
         sensitivity is analyzed.}
\label{tab:related_work}
\setlength{\tabcolsep}{5pt}
\renewcommand{\arraystretch}{1.4}
\small
\begin{tabular}{l ccc}
\toprule
\multirow{2}{*}{\textbf{Physics treatment}}
  & \multicolumn{3}{c}{\textbf{Sensitivity / data perturbation}} \\
\cmidrule(lr){2-4}
  & No sensitivity consideration
  & Training behavior
  & Post-hoc attribution \\
\midrule
No constraint
  & \makecell[l]{\citet{LIU2026132269}\\
                 \citet{MA2015187}\\
                 \citet{Fu7804912}\\
                 \citet{Li2017DiffusionCR}\\
                 \citet{ijcai2018p505}\\
                 \citet{Guo2019AttentionBS}\\
                 \citet{WANG2022127762}}
  & ---
  & \citet{koh17a} \\
\midrule
Soft constraint
  & \makecell[l]{\citet{MrquezNeila2017ImposingHC}\\
                 \citet{Dener2020TrainingNN}}
  & \makecell[l]{\citet{10.1016/j.jcp.2021.110768}\\
                 \citet{deryck2024}\\
                 \citet{pmlr-v202-daw23a}\\
                 \citet{wang2023expertsguidetrainingphysicsinformed}\\
                 \citet{Lu_2021}\\
                 \citet{10.1016/j.jcp.2019.109136}}
  & \citet{krasowski2026pinnfluenceinterpretingpinnsinfluence} \\
\midrule
\textbf{Hard constraint}
  & \makecell[l]{\citet{inproceedings}\\
                 \citet{huang2021differentiableprojectionconstraineddeep}\\
                 \citet{pmlr-v202-wang23at}\\
                 \citet{2023arXiv230708336}\\
                 \citet{articleKonstantinov}\\
                 \citet{chen2024physicsinformedneuralnetworkshard}\\
                 \citet{min2025hardnethardconstrainedneuralnetworks}}
  & ---
  & \textbf{This work} \\
\bottomrule
\end{tabular}
\end{table*}
\newpage
\section{Methodology}

\subsection{A Hard-Constrained Neural Network Formulation}
\label{sec:hard_constrained}


Consider a hard-constrained neural network architecture, in which an unconstrained neural-network prediction is passed through a differentiable enforcement layer before being used as the final model output. Let $\mathcal{X} \subseteq \mathbb{R}^{n_x}$ denote the input space
and let $f_\theta : \mathcal{X} \to \mathbb{R}^{n_y}$ be a neural
network parameterized by $\theta$, where $n_x$ is the input dimension
and $n_y$ is the output dimension. For an input $x \in \mathcal{X}$,
the raw neural-network prediction is denoted by
$\hat{y}_\theta(x) = f_\theta(x) \in \mathbb{R}^{n_y}$. In the
hard-constrained setting, this raw prediction is not directly used as
the final output because it may violate known constraints. Instead,
the final prediction is obtained by applying a projection operator $P$
to the raw output, yielding
\begin{equation}
    y^H_\theta(x) = P(f_\theta)(x),
    \label{eq:hard_prediction}
\end{equation}
where \( y^H_\theta(x) \) represents the hard-constrained prediction. The goal is to create an enforcement mapping \( P(f_\theta):\mathcal{X} \rightarrow \mathbb{R}^{n_\textup{y}} \) that meets the specified constraints for every \( x \in \mathcal{X} \).   Although the projection does not need to be globally smooth, being differentiable almost everywhere allows gradients to propagate through the projection layer, thereby enabling end-to-end gradient-based training \citep{min2025hardnethardconstrainedneuralnetworks}.  By construction, this hard-constrained model ensures that final output satisfies the constraints, unlike many physics-informed learning models that impose constraints softly through adding a penalty term to the training loss. 

We consider the HardNet-Aff formulation for enforcing input-dependent affine constraints \citep{min2025hardnethardconstrainedneuralnetworks}. Specifically, suppose the admissible output must satisfy
\begin{equation}
    b^\ell(x) \leq A(x)\,y \leq b^u(x), \quad \forall\, x \in
    \mathcal{X},
\end{equation}
where $A(x) \in \mathbb{R}^{n_c \times n_y}$ is the constraint matrix, $b^\ell(x), b^u(x) \in \mathbb{R}^{n_c}$ are the lower and upper constraint limits, and $n_c$ is the number of affine constraint
rows. In transportation networks, this class of constraints includes, for example, flow-conservation constraints at a node $i$,
\begin{equation}
\sum_{a\in\delta^-(i)} y_a
-
\sum_{a\in\delta^+(i)} y_a
= d_i,
\end{equation}
where $\delta^-(i)$ and $\delta^+(i)$ denote the sets of incoming and
outgoing links, respectively, and $d_i$ represents the net flow
supply or demand at node $i$. Such an equality can be represented in
the two-sided inequality form by setting the corresponding lower and
upper bounds equal, $b_i^\ell(x)=b_i^u(x)=d_i$. Similarly, a link-capacity constraint for link $a$ can be written as $0 \leq y_a \leq c_a(x)$. Where $c_a(x)$ denotes the available capacity of link $a$. Thus, both network flow-conservation and capacity constraints can be represented within the general affine constraint form considered here. Inequalities are understood component-wise. The feasible set
associated with input $x$ is therefore:
\begin{equation}
\mathcal{C}(x) = \left\{y \in \mathbb{R}^{n_y} :
b^\ell(x) \leq A(x)y \leq b^u(x)\right\}.
\end{equation}

The HardNet-Aff projection maps the raw neural-network output
$f_\theta(x)$ to a feasible point in $\mathcal{C}(x)$ through the
closed-form correction
\begin{equation}
    P(f_\theta)(x) = f_\theta(x) + A(x)^{+}\!\left[
    \mathrm{ReLU}\!\left(b^\ell(x) - A(x)f_\theta(x)\right) -
    \mathrm{ReLU}\!\left(A(x)f_\theta(x) - b^u(x)\right)\right],
    \label{eq:hardnet_aff}
\end{equation}
where $\mathrm{ReLU}(r) = \max\{0,r\}$ is applied component-wise and
\begin{equation}
    A(x)^{+} = A(x)^\top\!\left(A(x)A(x)^\top\right)^{-1}
\end{equation}
is the right pseudoinverse of $A(x)$. This expression assumes that
$A(x)$ has full row rank for each $x \in \mathcal{X}$, so that
$A(x)A(x)^\top$ is invertible, and that $\mathcal{C}(x)$ is nonempty
for every input.

\begin{figure}[t]
    \centering
    \includegraphics[width=0.95\textwidth]{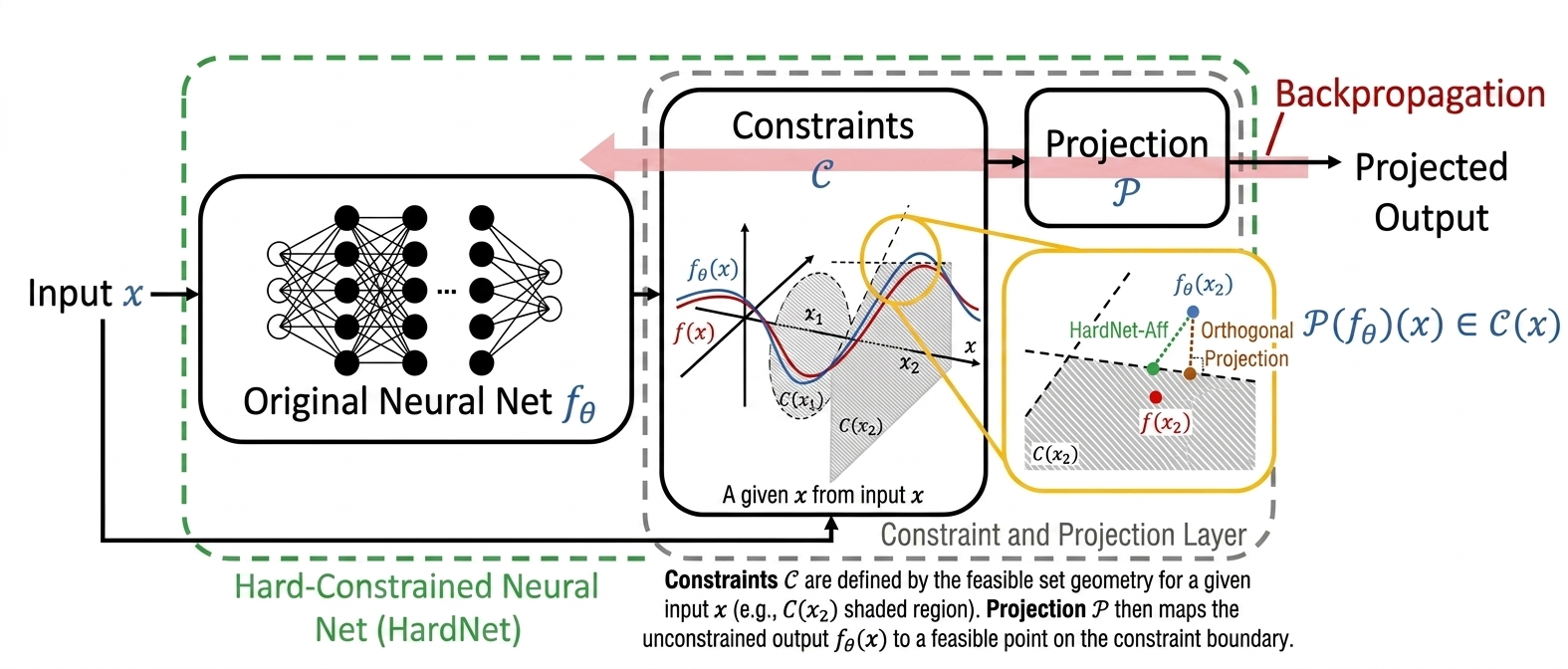}
    \caption{
    Schematic illustration of the HardNet-Aff framework, adapted from Min and Azizan~\cite{min2025hardnethardconstrainedneuralnetworks}. 
    A standard neural network first maps the input $x$ to an unconstrained prediction $f_\theta(x)$. 
    The prediction is then passed through a differentiable constraint-enforcement layer, where the affine constraint set $\mathcal{C}(x)$ is defined for the given input and the projection operator $\mathcal{P}$ maps the raw output to a feasible point. 
    The final projected output satisfies $\mathcal{P}(f_\theta)(x)\in\mathcal{C}(x)$ by construction, while gradients can still be backpropagated through the projection layer during end-to-end training.
    }
    \label{fig:hardnet_aff_framework}
\end{figure}

A useful special case arises when the constraints are equalities of
the form 
\begin{equation}
    A(x)y = b(x).
    \label{eq:equality}
\end{equation}
Unlike~\eqref{eq:hardnet_aff}, whose ReLU-based correction introduces non-differentiability, the following equality projection  in~\eqref{eq:equality_projection} is affine in $f_\theta(x)$ for fixed $x$ and therefore here smoothness is preserved. In this case,~\eqref{eq:hardnet_aff} simplifies to
\begin{equation}
    P(f_\theta)(x) = f_\theta(x) -
    A(x)^{+}\!\left(A(x)f_\theta(x) - b(x)\right),
    \label{eq:equality_projection}
\end{equation}
and under the full-row-rank assumption, $A(x)A(x)^{+} = I_{n_c}$,
so that the projected output satisfies $A(x)y^H_\theta(x) = b(x)$
exactly for every $x \in \mathcal{X}$. Even though this study does not deal with general nonlinear constraints, our formulation does extend the enforcement of hard constraints to cases involving input-dependent affine constraints with exact feasibility.
Figure~\ref {fig:hardnet_aff_framework} explains how the projection will be applied in our framework.  

\subsection{Hard Physics-Informed Influence Functions}
\label{sec:hard_if}

Influence functions are used to quantify how each training sample affects the
learned model and its physically feasible predictions. Let
$D = \{z_i = (x_i,\, y_i)\}_{i=1}^{N}$ be the training dataset,
where $x_i \in \mathcal{X}$ and $y_i \in \mathbb{R}^{m}$. Since the
final model output is $y^H_\theta(x)$, the sample-wise hard
physics-informed loss is defined as
\begin{equation}
    \ell^H(z_i;\theta) = \ell\!\left(y^H_\theta(x_i),\, y_i\right),
\end{equation}
where $\ell(\cdot,\cdot)$ is the prediction loss function. The empirical
hard-constrained risk and the trained parameters are
\begin{equation}
    L^H(\theta) = \frac{1}{N}\sum_{i=1}^{N}\ell^H(z_i;\theta)    
\end{equation}
To measure the contribution of training sample $z_i$, we consider an
infinitesimal upweighting of its loss term and define the perturbed
objective as
\begin{equation}
    L^H_{\epsilon,i}(\theta)
    =
    \frac{1}{N}\sum_{j=1}^{N}\ell^H(z_j;\theta)
    + \epsilon\,\ell^H(z_i;\theta).
\end{equation}
Henceforth, we assume that the relevant local minimizer is unique in a
neighborhood of $\hat{\theta}$, and denote the corresponding perturbed
solution by $\hat{\theta}_{\epsilon,i}$. The parameter-level influence is
the first-order change
$d\hat{\theta}_{\epsilon,i}/d\epsilon\big|_{\epsilon=0}$, approximated by
\begin{equation}
    \mathcal{I}^H_{\mathrm{param}}(z_i)
    =
    -H_H^{-1}\,\nabla_\theta\ell^H(z_i;\hat{\theta}),
    \qquad
    H_H
    =
    \nabla^2_\theta L^H(\theta)\big|_{\theta=\hat{\theta}},
    \label{eq:param_influence}
\end{equation}
where $H_H$ is the Hessian of the empirical hard-constrained risk evaluated at $\hat{\theta}$. This first-order approximation relies on the standard assumptions of the influence function that $L^{H}(\theta)$ is twice continuously differentiable in a neighborhood of $\hat{\theta}$, $H_H$ is non-singular, and the perturbation $\epsilon$ is infinitesimal; see, e.g., \cite{royset2021optimization} for general subdifferential calculus rules.

For a test sample $z_t = (x_t,\, y_t)$, the hard physics-informed
influence of training sample $z_i$ on the test loss at $z_t$ is
\begin{equation}
    \mathcal{I}^H(z_i,\, z_t) =
    -\nabla_\theta\ell^H(z_t;\hat{\theta})^\top
    H_H^{-1}\,\nabla_\theta\ell^H(z_i;\hat{\theta}).
    \label{eq:hard_if_score}
\end{equation}
Defining $g^H_i = \nabla_\theta\ell^H(z_i;\hat{\theta})$ and
$g^H_t = \nabla_\theta\ell^H(z_t;\hat{\theta})$, this becomes
$\mathcal{I}^H(z_i,z_t) = -(g^H_t)^\top H_H^{-1} g^H_i$. A positive
value indicates that upweighting $z_i$ increases the
hard-constrained test loss at $z_t$; a negative value indicates it
decreases it.

The main distinction from the classical influence function
\citep{koh17a} is that all derivatives are computed
through the hard-constrained prediction map. By the chain rule,
\begin{equation}
    \nabla_\theta\ell^H(z_i;\theta) =
    \frac{\partial\ell}{\partial y^H_\theta}
    \cdot\frac{\partial y^H_\theta}{\partial\hat{y}_\theta}
    \cdot\frac{\partial\hat{y}_\theta}{\partial\theta},
    \label{eq:chain_rule}
\end{equation}
so the Jacobian $\partial y^H_\theta/\partial\hat{y}_\theta$ includes the impact of the projection layer in all calculations of influence.
Consequently, $\mathcal{I}^H(z_i,z_t)$ measures the sensitivity of
the physically feasible prediction, not the unconstrained
neural-network output.

More generally, the same framework evaluates the influence of a
training sample on any differentiable scalar quantity
$\psi^H(x_t;\theta)$ derived from the hard-constrained prediction:
\begin{equation}
    \mathcal{I}^H_\psi(z_i,\,x_t) =
    -\nabla_\theta\psi^H(x_t;\hat{\theta})^\top
    H_H^{-1}\,\nabla_\theta\ell^H(z_i;\hat{\theta}).
    \label{eq:general_if}
\end{equation}
For a group of training samples $G \subseteq \{1,\ldots,N\}$, the
group influence on $z_t$ is
\begin{equation}
    \mathcal{I}^H(G,\,z_t) =
    -\left(g^H_t\right)^\top H_H^{-1}
    \sum_{i \in G} g^H_i.
    \label{eq:group_if}
\end{equation}
This group formulation supports sensitivity analysis over subsets of the training data such as spatial regions, time periods, operating conditions, or physical regimes.


\subsection{Null-Space Decomposition and the Physical Influence Purity Indicator}
\label{sec:pip}

For the equality constraints in~\eqref{eq:equality}, the
projection in~\eqref{eq:equality_projection} specializes to
\begin{equation}
    y^H_\theta(X_t)
    =
    \hat{y}_\theta(X_t)
    -
    A_c^+\!\left(A_c\hat{y}_\theta(X_t)-b_c\right).
    \label{eq:equality_projection_specific}
\end{equation}
Assuming that $A_c\in\mathbb{R}^{r\times m}$ has full row rank, where $m$ is the number of
monitored links and $r$ the number of conservation equations, its right pseudoinverse is $A_c^+=A_c^\top(A_cA_c^\top)^{-1}$. The corresponding projection Jacobian, obtained from~\eqref{eq:chain_rule}, is
\begin{equation}
    \frac{\partial y^H_\theta}{\partial\hat{y}_\theta}
    =
    I_m-A_c^+A_c
    =
    I_m-A_c^\top\!\left(A_cA_c^\top\right)^{-1}A_c
    =:\Pi_c .
    \label{eq:projector}
\end{equation}
Here, $\Pi_c\in\mathbb{R}^{m\times m}$ is the orthogonal projector onto $\mathrm{null}(A_c)$, representing directions that do not affect
the satisfaction of conservation constraints. Meanwhile, $I_m-\Pi_c=A_c^+A_c$ projects onto the constraint-related directions. This represents the component that is adjusted to enforce conservation
constraints.
The projector $\Pi_c$ is symmetric, idempotent, and depends only on the constraint matrix $A_c$; once the traffic network and conservation constraints are specified, $\Pi_c$ remains fixed during training.

Substituting Equation~\eqref{eq:projector} into Equation~\eqref {eq:chain_rule}, the gradient of the hard-constrained loss with respect to the raw prediction is
\begin{equation}
    \nabla_{\hat{y}}\ell^H(z_i;\hat\theta)
    =
    \Pi_c\,\delta_i,
    \qquad
    \delta_i
    :=
    \nabla_{y^H}
    \ell\!\left(y^H_{\hat{\theta}}(X_i),\,y_i\right)
    \in\mathbb{R}^m ,
    \label{eq:output_grad}
\end{equation}
where $z_i=(X_i,y_i)$ and $\delta_i$ is the gradient of the prediction loss
with respect to the hard-constrained output at training sample $z_i$. Since
$\Pi_c$ is an orthogonal projector, any $\delta_i\in\mathbb{R}^m$ admits the
unique orthogonal decomposition
\begin{equation}
    \delta_i
    =
    \underbrace{\Pi_c\delta_i}_{\delta_i^{\mathrm{phys}}}
    +
    \underbrace{(I_m-\Pi_c)\delta_i}_{\delta_i^{\mathrm{viol}}},
    \label{eq:decomposition}
\end{equation}
where $\delta_i^{\mathrm{phys}}$ is the conservation-consistent component and
$\delta_i^{\mathrm{viol}}$ is the conservation-violating component. This
interpretation follows directly from the geometry of the feasible set. If
$y$ is feasible to satisfy $A_cy=b_c$, for any
$v\in\mathrm{null}(A_c)$ we have $A_c(y+\eta v)=b_c$
 for all $\eta\in\mathbb{R}$. Thus, moving in a null-space direction does not affect conservation. For a nonzero $v\in\mathrm{row}(A_c)$, we can
write $v=A_c^\top w$. Since $A_c$ has full row rank,
$A_cA_c^\top$ is nonsingular, and therefore $A_cv\neq0$. Hence, for $\eta\neq0$, moving in this direction changes the constraint residual.
The decomposition in~\eqref{eq:decomposition} therefore separates the gradient into directions that are compatible with the conservation
constraints and directions that are removed by the projection. The orthogonality of these two components yields
\begin{equation}
    \|\delta_i\|_2^2
    =
    \|\delta_i^{\mathrm{phys}}\|_2^2
    +
    \|\delta_i^{\mathrm{viol}}\|_2^2 .
    \label{eq:pythagorean}
\end{equation}

Equation \eqref{eq:output_grad} shows that only $\delta_i^{\mathrm{phys}}$ contributes to the gradient with respect to the raw prediction. The component $\delta_i^{\mathrm{viol}}$ is in the row space of $A_c$ and is eliminated by the projection Jacobian $\Pi_c$. Therefore, under the projected loss, this component does not propagate to the model parameters through the hard-constrained prediction map.

The decomposition also shows how much of a sample gradient is retained after the constraint projection. To compare this effect across samples, the measure should reflect the gradient's alignment with $\mathrm{null}(A_c)$ rather than
its magnitude; otherwise, samples with larger residuals would dominate the comparison. A common range is also useful when comparing different time
periods or network locations. Since $\Pi_c$ acts on all $m$ output components at once, the measure should be based on the full gradient vector.

The orthogonal decomposition in~\eqref{eq:pythagorean} gives this quantity
directly. So the retained component can be normalized by the total gradient norm.
The resulting ratio is independent of gradient magnitude, lies in $[0,1]$, and measures the share of the squared gradient norm that is aligned with
conservation-consistent directions.

\begin{definition}[Physical Influence Purity]
The Physical Influence Purity (PIP) of training sample $z_i$ is defined as
\begin{equation}
    \mathrm{PIP}(z_i)
    =
    \frac{\|\delta_i^{\mathrm{phys}}\|_2^2}
    {\|\delta_i\|_2^2+\varepsilon}
    =
    \frac{\|\Pi_c\delta_i\|_2^2}
    {\|\delta_i\|_2^2+\varepsilon},
    \label{eq:pip}
\end{equation}
where $\varepsilon>0$ is a small constant used to avoid division by zero.
When $\|\delta_i\|_2>0$, $\mathrm{PIP}(z_i)\in[0,1]$.
\end{definition}

PIP is bounded in $[0,1]$ directly from the orthogonal decomposition in~\eqref{eq:pythagorean}. In particular, both
$\|\delta_i^{\mathrm{phys}}\|_2^2$ and
$\|\delta_i^{\mathrm{viol}}\|_2^2$ are non-negative.
Therefore, $\|\delta_i^{\mathrm{phys}}\|_2^2$ cannot exceed
$\|\delta_i\|_2^2$, and the ratio in~\eqref{eq:pip} lies between
zero and one. Squared norms are used because~\eqref{eq:pythagorean} gives an additive decomposition of the squared gradient magnitude. Thus, PIP represents the fraction of the total
squared gradient magnitude that remains in conservation-consistent directions after projection.

A value close to $\mathrm{PIP}(z_i)=1$ means that the loss-gradient direction
is nearly aligned with $\mathrm{null}(A_c)$. Thus, the change indicated by
the gradient is largely compatible with the conservation constraints. In contrast, a value close to $\mathrm{PIP}(z_i)=0$ means that
the gradient lies mainly in $\mathrm{row}(A_c)$. This component is removed when the gradient is multiplied by the projection matrix $\Pi_c$ and
therefore does not contribute to updates through the prediction.

\subsection{Practical Computation of Influence Functions}
\label{sec:practical_computation}

The influence score in Equation~\eqref{eq:hard_if_score}
requires computing $H_H^{-1}\mathbf{g}_t^H$, where
$H_H = \nabla^2_\theta L^H(\hat\theta) \in \mathbb{R}^{p \times p}$
is the Hessian of the hard-constrained empirical loss. With a high number of trainable parameters,
forming $H_H$ is computationally infeasible, and direct inversion costs $O(p^3)$ operations.
The key observation is that we never need $H_H^{-1}$ as
an explicit matrix.
For a fixed test window $z_t$, every influence score
$\mathcal{I}^H(z_i, z_t) = -s_t^\top \mathbf{g}_i^H$
reduces to a dot product once the
inverse Hessian-vector product (iHVP)
\begin{equation}
  s_t = H_H^{-1}\,\mathbf{g}_t^H
  \label{eq:ihvp_def}
\end{equation}
is available.
The computational task therefore reduces to solving the linear system $H_H s_t = \mathbf{g}_t^H$ without ever forming $H_H$.

The iterative solver we adopt requires only Hessian-vector products (HVPs),
that is, products of the form $H_Hv$ for arbitrary
$v\in\mathbb{R}^p$.
By the Pearlmutter identity~\citep{pearlmutter1994fast}, this product is equal to a gradient:
\begin{equation}
  H_H\,v
  = \nabla_\theta\!\left(\nabla_\theta L^H(\hat\theta)^\top v\right).
  \label{eq:hvp}
\end{equation}
The right-hand side is the gradient of the scalar
$\nabla_\theta L^H \cdot v$ with respect to $\theta$.
This costs $O(p)$ time and memory with no need to store $H_H$.
A significant structural outcome is that every
Hessian-vector product is automatically physics-informed.
By the chain rule, the gradient of the hard-constrained
loss satisfies:
\begin{equation}
  \mathbf{g}^H
  = \nabla_\theta \ell^H
  = \frac{\partial \ell}{\partial y^H}
    \cdot \Pi_c \cdot
    \frac{\partial \hat{y}}{\partial \theta},
  \label{eq:hard_gradient}
\end{equation}
so the conservation projection $\Pi_c$ acts on every gradient
before it enters the HVP. Differentiating~\eqref{eq:hard_gradient} again, the Hessian separates into a Gauss-Newton term of the form $J^\top\Pi_c\bigl(\nabla^2_{y^H}\ell\bigr)\Pi_cJ$. Here, $J=\partial\hat{y}/\partial\theta$, and there is also a term that involves the second derivative of the network output that is weighted by $\Pi_c\delta$. In the Gauss-Newton term, any change in parameters that would push the raw prediction in a direction that violates conservation is canceled by $\Pi_c$ and therefore does not contribute to the curvature term.

To approximate $s_t$ iteratively, we rely on the
Neumann series identity.
For any step size $\alpha$ satisfying
$0 < \alpha < 2/\rho(H_H)$,
the inverse admits the convergent series:
\begin{equation}
  H_H^{-1}
  = \alpha \sum_{k=0}^{\infty}(I - \alpha H_H)^k,
  \label{eq:neumann}
\end{equation}
which holds because all eigenvalues of $(I - \alpha H_H)$
are strictly inside the unit circle under the stated
condition on $\alpha$.
Multiplying both sides of~\eqref{eq:neumann} by $\mathbf{g}_t^H$ and
identifying the partial sums with an iterative update leads to the
LiSSA recursion~\citep{Lissa}:
\begin{equation}
  h_0 = \alpha\,\mathbf{g}_t^H,
  \qquad
  h_{k+1} = \alpha\,\mathbf{g}_t^H + h_k - \alpha\,H_H\,h_k,
  \label{eq:lissa_basic}
\end{equation}
where $h_K \to s_t$ exactly as $K \to \infty$.
Each step requires exactly one HVP evaluation.

For our model, however, the na\"{i}ve
recursion~\eqref{eq:lissa_basic} converges very slowly in practice.
A large spectral radius $\rho(H_H)$, restricts the admissible step size $\alpha$, while
near-zero Hessian eigenvalues lead to poor conditioning. The solution for this problem is to normalize $H_H$ by its spectral radius before
applying the Neumann series.
Define the scaled and damped Hessian:
\begin{equation}
  \widetilde{H}_d
  = \frac{H_H}{\rho(H_H)} + dI,
  \label{eq:scaled_hessian}
\end{equation}
where $d > 0$ is a small damping constant.
Dividing by $\rho(H_H)$ maps all eigenvalues of $H_H$
into $[0,1]$, so $\widetilde{H}_d$ has eigenvalues in
$[d, 1+d]$ and is positive definite.
The Neumann series for $\widetilde{H}_d^{-1}$ then converges
with the step size $\alpha = 1$, which yields the update rule:
\begin{equation}
  h_0 = \mathbf{g}_t^H,
  \qquad
  h_{k+1}
  = \mathbf{g}_t^H
    + (1-d)\,h_k
    - \frac{1}{\rho(H_H)}\,H_H\,h_k.
  \label{eq:kl_update}
\end{equation}
If $\lambda_i$ denotes the $i$-th eigenvalue of $H_H$,
the corresponding eigenvalue of the iteration matrix
$(I - \widetilde{H}_d)$ is:
\begin{equation}
  \mu_i
  = 1 - \frac{\lambda_i}{\rho(H_H)} - d
  \;\in\; [-d,\; 1-d],
  \label{eq:iter_eigenvalues}
\end{equation}
where we assume $H_H \succeq 0$, so that $\lambda_i \in [0,\rho(H_H)]$ and hence $\lambda_i/\rho(H_H) \in [0,1]$.
The spectral radius of the iteration matrix is therefore:
\begin{equation}
  \rho(I - \widetilde{H}_d)
  = \max_i |\mu_i|
  = \max(d,\; 1-d)
  = 1 - d.
  \label{eq:iter_radius}
\end{equation}

\begin{theorem}[Convergence of Scaled LiSSA]
\label{thm:lissa_convergence}
Let $h^* = \widetilde{H}_d^{-1}\,\mathbf{g}_t^H$ denote
the unique solution to the scaled linear system
$\widetilde{H}_d\,h = \mathbf{g}_t^H$.
Starting from $h_0 = \mathbf{g}_t^H$, the
iterate $h_K$ produced by~\eqref{eq:kl_update} satisfies:
\begin{equation}
  \|h_K - h^*\| \leq (1-d)^K\,\|h_0 - h^*\|.
  \label{eq:convergence_bound}
\end{equation}
The rate $(1-d)^K$ depends only on $d$ and $K$,
not on $\rho(H_H)$ or $\kappa(H_H)$.
\end{theorem}

\begin{proof}{Proof.}
\textbf{Step 1: Identify the fixed point.}
The update rule~\eqref{eq:kl_update} is a linear map in $h_k$.
A fixed point $h^*$ satisfies $h_{k+1} = h_k = h^*$,
so substituting into~\eqref{eq:kl_update}:
\begin{equation}
  h^*
  = \mathbf{g}_t^H + (1-d)\,h^*
    - \frac{1}{\rho(H_H)}\,H_H\,h^*.
  \label{eq:fp_raw}
\end{equation}
Collecting all terms involving $h^*$ on the left:
\begin{equation}
  h^* - (1-d)\,h^* + \frac{1}{\rho(H_H)}\,H_H\,h^*
  = \mathbf{g}_t^H.
  \label{eq:fp_collect}
\end{equation}
Simplifying the left side:
\begin{equation}
  d\,h^* + \frac{H_H}{\rho(H_H)}\,h^*
  = \mathbf{g}_t^H.
  \label{eq:fp_simplified}
\end{equation}
Factoring out $h^*$ and recognizing the
definition~\eqref{eq:scaled_hessian}:
\begin{equation}
  \underbrace{\left(\frac{H_H}{\rho(H_H)} + dI\right)}_{\widetilde{H}_d}
  h^* = \mathbf{g}_t^H.
  \label{eq:fp_final}
\end{equation}
Since $\widetilde{H}_d$ is positive definite it is invertible,
and the unique fixed point is:
\begin{equation}
  h^* = \widetilde{H}_d^{-1}\,\mathbf{g}_t^H.
  \label{eq:fixed_point_solution}
\end{equation}

\textbf{Step 2: Derive the error recursion.}
The update rule~\eqref{eq:kl_update} can be written compactly as:
\begin{equation}
  h_{k+1} = \mathbf{g}_t^H + (I - \widetilde{H}_d)\,h_k.
  \label{eq:update_compact}
\end{equation}
The fixed point equation~\eqref{eq:fp_final} equivalently reads:
\begin{equation}
  h^* = \mathbf{g}_t^H + (I - \widetilde{H}_d)\,h^*.
  \label{eq:fp_compact}
\end{equation}
Define the error $e_k = h_k - h^*$.
Subtracting~\eqref{eq:fp_compact} from~\eqref{eq:update_compact}:
\begin{align}
  e_{k+1}
  &= h_{k+1} - h^* \notag\\
  &= \bigl[\mathbf{g}_t^H + (I - \widetilde{H}_d)\,h_k\bigr]
     - \bigl[\mathbf{g}_t^H + (I - \widetilde{H}_d)\,h^*\bigr]
     \notag\\
  &= (I - \widetilde{H}_d)\,(h_k - h^*)
  \notag\\
  &= (I - \widetilde{H}_d)\,e_k.
  \label{eq:error_recursion}
\end{align}

\textbf{Step 3: Apply the recursion $K$ times.}
Equation~\eqref{eq:error_recursion} shows that the error is
multiplied by $(I - \widetilde{H}_d)$ at every step.
Applying it repeatedly:
\begin{equation}
  e_K = (I - \widetilde{H}_d)^K\,e_0.
  \label{eq:error_power}
\end{equation}

\textbf{Step 4: Bound the norm.}
Since $H_H$ is symmetric and $I$ is symmetric,
the iteration matrix $I - \widetilde{H}_d$ is symmetric, hence
normal, so its spectral norm equals its spectral radius:
$\|I-\widetilde{H}_d\|_2 = \rho(I-\widetilde{H}_d)$.
Taking the Euclidean norm of both sides
of~\eqref{eq:error_power} and applying this fact together with
$\rho(A^K) = \rho(A)^K$:
\begin{equation}
  \|e_K\|
  = \|(I - \widetilde{H}_d)^K\,e_0\|
  \leq \rho\!\left((I - \widetilde{H}_d)^K\right)\|e_0\|
  = \rho(I - \widetilde{H}_d)^K\,\|e_0\|.
  \label{eq:norm_bound}
\end{equation}
Substituting~\eqref{eq:iter_radius}:
\begin{equation}
  \|h_K - h^*\|
  \leq (1 - d)^K\,\|h_0 - h^*\|. \tag*{\(\square\)}
  \label{eq:final_bound}
\end{equation}
\end{proof}

It is worth being precise about what the scaled LiSSA
computes relative to the exact iHVP.
From equation~\eqref{eq:fp_final}, the fixed point $h^*$
solves $\widetilde{H}_d\,h^* = \mathbf{g}_t^H$.
Substituting the definition~\eqref{eq:scaled_hessian}
and multiplying both sides by $\rho(H_H)$:
\begin{equation}
  \left(\frac{H_H}{\rho(H_H)} + dI\right)h^*
  = \mathbf{g}_t^H
  \;\;\Longrightarrow\;\;
  \bigl(H_H + \lambda I\bigr)\,h^*
  = \rho(H_H)\,\mathbf{g}_t^H,
  \label{eq:regularised_system}
\end{equation}
where $\lambda = d\cdot\rho(H_H)$ is the effective
regularization.
Solving for $h^*$ gives
$h^* = \rho(H_H)\,(H_H + \lambda I)^{-1}\mathbf{g}_t^H$.
Substituting into the influence score:
\begin{equation}
  \mathcal{I}^H_\lambda(z_i, z_t)
  = -(\mathbf{g}_t^H)^\top
    (H_H + \lambda I)^{-1}
    \mathbf{g}_i^H
  = -\frac{(h^*)^\top \mathbf{g}_i^H}{\rho(H_H)}.
  \label{eq:regularised_if}
\end{equation}
This is the regularized influence function, which replaces
the exact $H_H^{-1}$ with $(H_H + \lambda I)^{-1}$.
Convergence is true for any $d>0$ according to Theorem~\ref{thm:lissa_convergence}. The regularization $\lambda = d\cdot\rho(H_H)$ reflects the cost of a contraction rate that does not depend on the conditioning of $H_H$. This regularization can be made tiny, but this will require more iterations.
Since $\rho(H_H)$ is a global constant shared across all
pairs $(z_i, z_t)$, it cancels upon normalization and
does not affect the relative ranking of training samples.

\subsection{Station Sensitivity Score}
\label{sec:sss}

The influence score $\mathcal{I}^H(z_i, z_t)$ ranks training samples. To
measure how sensitive the forecast at each detector station is, we define a
score for each station that uses the $s_t$
from~\eqref{eq:ihvp_def}. For a test window $z_t$ with input $X_t$, let
\begin{equation}
  \hat{s}_t = \frac{s_t}{\lVert s_t \rVert}
  \label{eq:unit_direction}
\end{equation}
be the unit direction of the iHVP, and let $y^H_e(X_t;\hat\theta)$ be the
physics-feasible forecast at station $e \in \{1,\dots,N\}$. The Station
Sensitivity Score (SSS) of station $e$ is
\begin{equation}
  \mathrm{SSS}_e
  = \frac{1}{|\mathcal{T}|}\sum_{z_t \in \mathcal{T}}
    \left|\, \hat{s}_t^{\top}\, \nabla_\theta\, y^H_e(X_t;\hat\theta) \,\right|,
  \label{eq:sss}
\end{equation}
where $\mathcal{T}$ is the set of test windows. The term inside the absolute
value is the rate of change of the forecast at station $e$ when the parameters
move along $\hat{s}_t$. A large $\mathrm{SSS}_e$ means that the forecast at
station $e$ responds strongly to the parameter changes that matter most for
the test windows. For comparison across stations, the scores are scaled.


\section{Numerical Experiments}

\subsection{Traffic Flow Prediction Problem Setup}
\label{sec:traffic_model}

We formulate traffic flow forecasting on a directed traffic network
$\mathcal{G} = (\mathcal{V},\mathcal{E})$, where $\mathcal{V}$ is the
set of junctions and $\mathcal{E} = \{1,\ldots,m\}$ is the set of
monitored directed links. At time $t$, the feature vector of link
$e \in \mathcal{E}$ is $s_e(t) = [q_e(t),\,v_e(t),\,o_e(t)]^\top$,
where $q_e(t)$, $v_e(t)$, and $o_e(t)$ denote flow, speed, and
occupancy, respectively. Stacking all link-level features gives the
network traffic-state matrix $S(t) \in \mathbb{R}^{m \times 3}$.

To capture temporal dependence, we use a rolling time window of
length $w$. The input sample at time $t$ is
$X_t = \{S(t-w+1),\ldots,S(t)\} \in \mathbb{R}^{w \times m \times 3}$,
and the prediction target is the next-hour flow vector
$y_t = [q_1(t+1),\ldots,q_m(t+1)]^\top \in \mathbb{R}^m$.
The raw neural-network prediction is
$\hat{y}_\theta(X_t) = f_\theta(X_t)$, where
$f_\theta : \mathbb{R}^{w \times m \times 3} \to \mathbb{R}^m$ is a
spatiotemporal forecasting model, and the final prediction is the
hard-constrained output $y^H_\theta(X_t)$.

The physical constraint considered in this work is junction-level flow
conservation. Let $\mathcal{V}_c \subseteq \mathcal{V}$ denote the
set of junctions where conservation is imposed. For each
$u \in \mathcal{V}_c$, let $\mathcal{I}(u) \subseteq \mathcal{E}$
and $\mathcal{O}(u) \subseteq \mathcal{E}$ be the sets of incoming
and outgoing monitored links. The predicted flows are required to
satisfy
\begin{equation}
    \sum_{e \in \mathcal{I}(u)} y_e
    - \sum_{e \in \mathcal{O}(u)} y_e = b_{c,u},
\end{equation}
where $b_{c,u}$ is the net imbalance at junction $u$. For example, if the time window is sufficient to clear the traffic, $b$ could be zero. Stacking over all constrained
junctions gives the equality constraint
\begin{equation}
    A_c\,y = b_c,
    \label{eq:conservation}
\end{equation}
where $A_c \in \mathbb{R}^{|\mathcal{V}_c| \times m}$ is the
junction--link conservation matrix with entries
\begin{equation}
    (A_c)_{u,e} =
    \begin{cases}
        \phantom{-}1, & \text{if link } e \text{ enters junction } u, \\
        -1, & \text{if link } e \text{ leaves junction } u, \\
        \phantom{-}0, & \text{otherwise,}
    \end{cases}
\end{equation}
and $b_c \in \mathbb{R}^{|\mathcal{V}_c|}$ is the vector of net
junction imbalances. Using the HardNet-Aff projection from
Section~\ref{sec:hard_constrained}, the final forecast satisfies
$A_c\,y^H_\theta(X_t) = b_c$ for every input window $X_t$.

Given the training dataset $D = \{(X_t,\,y_t)\}_{t=w}^{T-1}$, the
model is trained by minimizing
\begin{equation}
    \min_\theta\;\frac{1}{T-w}\sum_{t=w}^{T-1}
    \ell\!\left(y^H_\theta(X_t),\,y_t\right),
    \label{eq:training_objective}
\end{equation}
where $\ell(\cdot,\cdot)$ is the forecasting loss function.

\subsection{Experimental Setup}
We use hourly PeMS District~3 data covering 318 monitored
freeway links across the Sacramento metropolitan network from January to May 2025, yielding $T = 3{,}623$ hourly
observations. A map of the study area is provided in the Results section (Figure \ref{fig:sss_map}). Each link carries three features - flow,
speed, and occupancy - and the prediction target is the
next-hour flow vector $y_t \in \mathbb{R}^{318}$. Using a
24-hour input window, the dataset produces 3{,}599
input-target pairs, split chronologically into 70\%
training, 15\% validation, and 15\% test. Features are
standardized using training-set statistics; targets are
retained in original units (vehicles per hour).


We construct a node-link graph from corridor topology,
generating 312 candidate junction flow conservation
equations. Each equation is validated empirically: a
constraint is retained only if its mean relative residual
does not exceed 10\% and its 95th-percentile relative
residual does not exceed 30\% of total involved flow, with
all constituent links observed. Of the 312 candidates, 62
pass both criteria and form the constraint matrix
$A_c \in \mathbb{R}^{62 \times 318}$, which has full row
rank. Every retained constraint relates exactly one
consecutive station pair ($+1$ incoming, $-1$ outgoing).
The constraint vector $b_c$ is estimated as the mean of
$A_c y_{\mathrm{train}}(t)$ over the training period,
accounting for systematic junction imbalances due to
residual unmonitored flows.

\begin{figure}[h]
    \centering
    \includegraphics[width=0.95\linewidth]{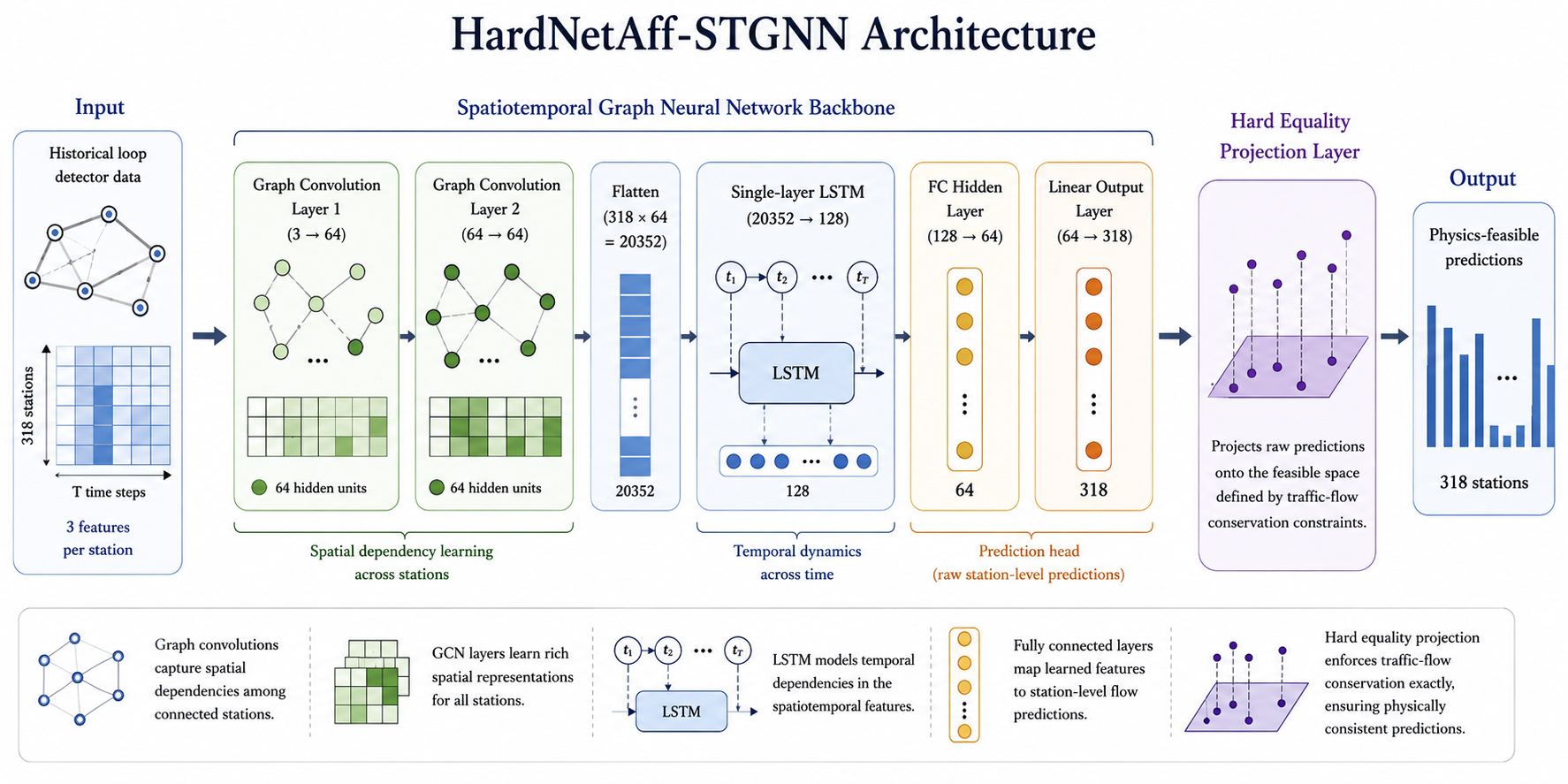}
    \caption{HardNetAff-STGNN architecture for traffic-flow prediction. The STGNN backbone learns spatial and temporal traffic patterns from loop detector data, while the hard equality projection layer enforces traffic-flow conservation to produce physics-feasible station-level predictions (figure generated in ChatGPT).}
    \label{fig:HardNetAff-STGNN_architecture}
\end{figure}

As shown in Figure \ref{fig:HardNetAff-STGNN_architecture}, HardNetAff-STGNN consists of a backbone of the spatial-temporal graph neural network followed by a hard equality projection layer. With approximately $10.5\times10^{6}$ trainable parameters, The backbone includes two graph convolutional layers with 64 hidden units, a single-layer LSTM with 128 hidden units, and a prediction head composed of a fully connected hidden layer with 64 units and a final linear output layer producing predictions for 318 stations. ReLU activation and dropout with a rate of 0.1 are used for nonlinearity and regularization. The raw station-level predictions are then passed through the hard affine equality projection layer to enforce traffic-flow conservation constraints exactly.
The model is trained end to end with the Adam
optimizer using a learning rate of $10^{-3}$, weight decay of $10^{-5}$, and a batch size of 32, with gradient clipping at a maximum norm of 5.0.

The spectral radius $\rho(H_H)$ is estimated by stochastic power
iteration.
For the main iHVP computation we use damping $d = 0.05$,
$K = 400$ iterations, and $J = 3$ independent chains, yielding a
deterministic convergence error of
$(0.95)^{400} \approx 1.2 \times 10^{-9}$ per
Theorem~\ref{thm:lissa_convergence}.
The batch size is set to $B = 256$, which reduces the stochastic
noise floor by a factor of $256/32 = 8$ compared to the commonly
used default of $B = 32$, bringing
run-to-run agreement below the $8\%$ threshold.
Convergence of each iHVP estimate is verified by the
run-agreement criterion:
\begin{equation}
  \frac{\|h^{(1)} - h^{(2)}\|}
       {\tfrac{1}{2}\|h^{(1)} + h^{(2)}\|}
  < 0.08,
  \label{eq:run_agreement}
\end{equation}
where $h^{(1)}$ and $h^{(2)}$ are the two independent chains
initialized from the same $\mathbf{g}_t^H$ but with different
random batch sequences.
Table~\ref{tab:lissa_params} summarizes all hyperparameters.
 
\begin{table}[h]
\centering
\caption{Scaled LiSSA hyperparameters used in experiments.}
\label{tab:lissa_params}
\begin{tabular}{p{3.2cm} c p{7.5cm}}
\toprule
Parameter & Value & Justification \\
\midrule
Damping $d$ & $0.05$
  & Convergence error $(0.95)^{400}
    \approx 1.2\times10^{-9}$
    (Theorem~\ref{thm:lissa_convergence}) \\
Iterations $K$ & $400$
  & Error bound  \\
Runs $J$ & $3$
  & Variance reduction; agreement verified
    by~\eqref{eq:run_agreement} \\
Batch size $B$ & $256$
  & Noise floor $\propto 1/B$;
    $8\times$ reduction vs.\ $B{=}32$;
    agreement $\approx 3$--$4\%$ \\
Power steps $T_\rho$ & $40$
  & Stable $\hat\rho$ estimate \\
Power batches $M$ & $12$
  & Accurate HVP per power step \\
Agreement threshold & $8\%$
  & Accounts for irreducible stochastic
    noise at $B{=}256$ \\
\bottomrule
\end{tabular}
\end{table}

All experiments were conducted in Google Colab using high-RAM runtime with a single NVIDIA T4 GPU.

\section{Results}

Table~\ref{tab:test_results} shows test set performance of
constrained and unconstrained models, measured by data alignment and constraint alignment. Data alignment is measured by

\begin{equation}
  \mathrm{MAE}
  = \frac{1}{T_{\mathrm{test}}\,m}
    \sum_{t,e}\bigl|\hat{y}_{t,e}-y_{t,e}\bigr|,
  \qquad
  \mathrm{RMSE}
  = \sqrt{\frac{1}{T_{\mathrm{test}}\,m}
    \sum_{t,e}\bigl(\hat{y}_{t,e}-y_{t,e}\bigr)^{2}},
  \label{eq:metrics}
\end{equation}
where $y_{t,e}$ and $\hat{y}_{t,e}$ are the observed and
predicted flow at link $e$ in test window $t$, both in
vehicles per hour on the original scale. The HardNetAff-STGNN and unconstrained model achieve a similar level of data alignment. The key difference is in the alignment with physical constraints. The unconstrained output breaks junction flow conservation by an average of 121.95~veh/hr, reaching as high as 536.143~veh/hr in some cases. This is equivalent to the hourly capacity of a full freeway lane. The hard-constrained output reduces this to $9.9 \times 10^{-5}$~veh/hr on average. This confirms near-exact enforcement of all 62 constraints at once throughout the test period.

\begin{table}[htbp]
\centering
\caption{Test set results on 318-link Sacramento PeMS network (540 test windows). Flow units: veh/hr.}
\label{tab:test_results}
\begin{tabular}{lccccc}
\toprule
 & \multicolumn{3}{c}{Data alignment} & \multicolumn{2}{c}{Physical constraint alignment} \\
\cmidrule(lr){2-4} \cmidrule(lr){5-6}
Model & MAE & RMSE & $R^2$ & $\overline{|Ay-b|}$ & $\max|Ay-b|$ \\
\midrule
Unconstrained    & 267.209 & 443.807 & 0.9147 & 121.95 & 536.143 \\
HardNetAff-STGNN & \textbf{263.659} & \textbf{439.798} & \textbf{0.9162} & $\mathbf{9.9\times10^{-5}}$ & $\mathbf{5.66\times10^{-4}}$ \\
\bottomrule
\end{tabular}
\end{table}

The small accuracy gain shows a common feature of hard-constrained architectures. The projection ensures physical feasibility by design instead of improving the raw predictive signal. Therefore, accuracy gains are less important than meeting constraint satisfaction guarantees.

\subsection{Physics-Informed Purity Analysis}
\label{sec:pip_results}

Table~\ref{tab:pip_summary} shows the PIP statistics for the HardNetAff-STGNN across the training sets.  These values represent the fraction of loss-gradient energy that aligns with the conservation-consistent subspace $\mathrm{null}(A_c)$. The HardNetAff-STGNN has a mean PIP of $0.923$ on the training set. The close match between training and test PIP indicates that the model's gradient structure for conservation performs well with new observations. A mean PIP of $0.923$ means that about $92.3\%$ of the loss-gradient energy at the hard-constrained output is in the conservation-consistent subspace $\mathrm{null}(A_c)$. The remaining $7.7\%$ is in the conservation-violating direction $\mathrm{row}(A_c)$, and the projection Jacobian $\Pi_c$ removes it before it affects the model parameters.

\begin{table}[ht]
\centering
\caption{PIP statistics for the HardNetAff-STGNN.}
\label{tab:pip_summary}
\begin{tabular}{lcc}
\toprule
& Training set  \\
\midrule
Mean PIP   & 0.923  \\
Median PIP & 0.931   \\
\midrule
AM peak  (07:00-09:00) & 0.948 \\
PM peak  (16:00-18:00) & 0.938  \\
Night    (22:00-05:00) & 0.891  \\
\bottomrule
\end{tabular}
\end{table}

PIP changes during different traffic periods. The AM peak period has the highest PIP at \$0.948. This indicates that morning commute patterns create loss gradients that closely match the conservation-consistent subspace. The PM peak has a PIP of \$0.938. The night period, from 22:00 to 05:00, shows the lowest PIP at \$0.891, which is \$0.057 lower than the AM peak. During overnight hours, low and highly variable traffic flow leads to larger prediction errors. This makes the physical constraints more influential.

\subsection{Station Sensitivity Score and Spatial Analysis}
\label{sec:sss_results}

\begin{figure}[ht]
   \centering
   \includegraphics[width=0.75\linewidth]{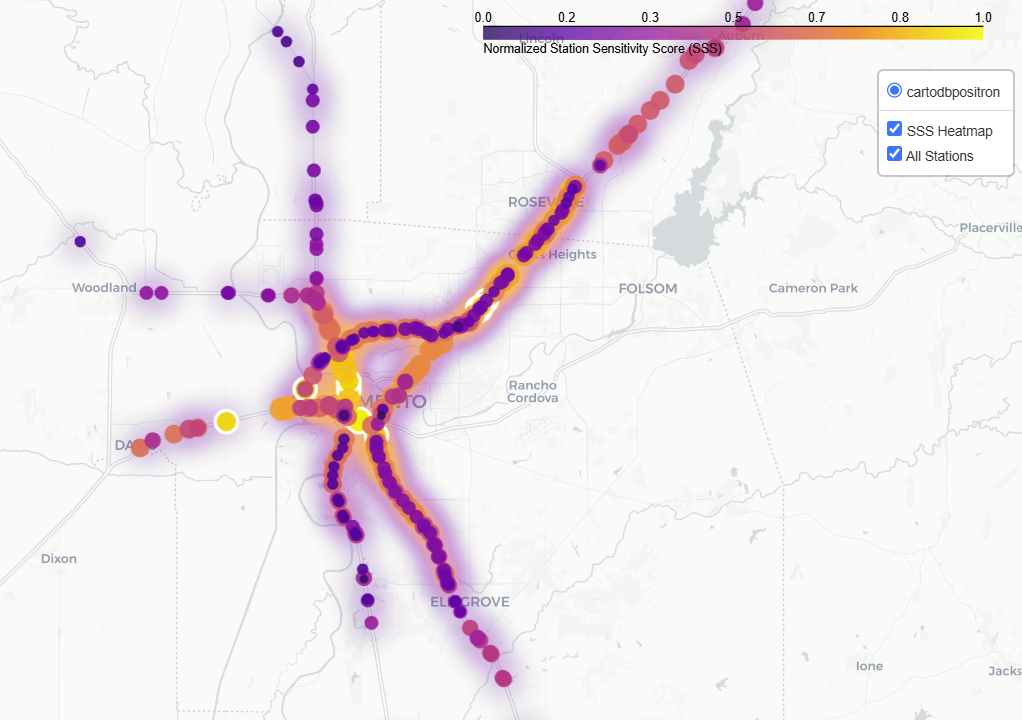}
   \caption{Spatial distribution of Station Sensitivity Scores
   (SSS) across the Sacramento PeMS network. Yellow indicates
   high sensitivity; dark blue indicates near-zero sensitivity.
   The star-shaped pattern centers on the downtown
   interchange where I-80, I-5, US-50, and CA-99 meet.}
   \label{fig:sss_map}
 \end{figure}

Figure~\ref{fig:sss_map} show how SSS values are spread across the 318 monitored freeway stations. The map highlights a clear star-shaped concentration around the downtown Sacramento interchange (lat~$\approx 38.6^\circ$, lon~$\approx -121.4^\circ$), where Interstates 80 and 5, US Highway 50, and State Route 99 meet. SSS declines steadily from this core along each corridor, dropping to nearly zero at outer stations more than about 50~km from the interchange. This pattern reflects the layout of the conservation constraint matrix $A_c$: stations near the most constrained junctions show the highest alignment with the network-level loss gradient $\hat{g}_t$. This finding indicates that the model's predictions for all 318 links depend most critically on the flows at these key nodes. The northeast arm, which corresponds to the inbound I-80 corridor from Roseville, remains elevated over the largest distance. This is consistent with this route carrying the highest inbound traffic volume and giving the most valuable gradient signal during training.

\subsection{Impact of Physics Constraints on Influence Assessment}

We compare three frameworks that differ in their models and influence methods. In the first framework, we apply the classic influence function to the unconstrained model, so no part of the analysis uses physics. In the second, we apply the classic influence function to the physics-informed model. The model is trained with our HardNet-Aff model, but the influence analysis ignores the projection layer and uses only the network's raw prediction. In the third, we apply our physics-informed influence function framework to the physics-informed model, so both the model and the influence analysis include the conservation constraints. Difference between the first and second frameworks shows the effect of considering physics in the model. Difference between the second and third shows the effect of considering physics in the influence analysis.

\begin{figure}[htbp]
    \centering
    \includegraphics[width=0.75\textwidth]{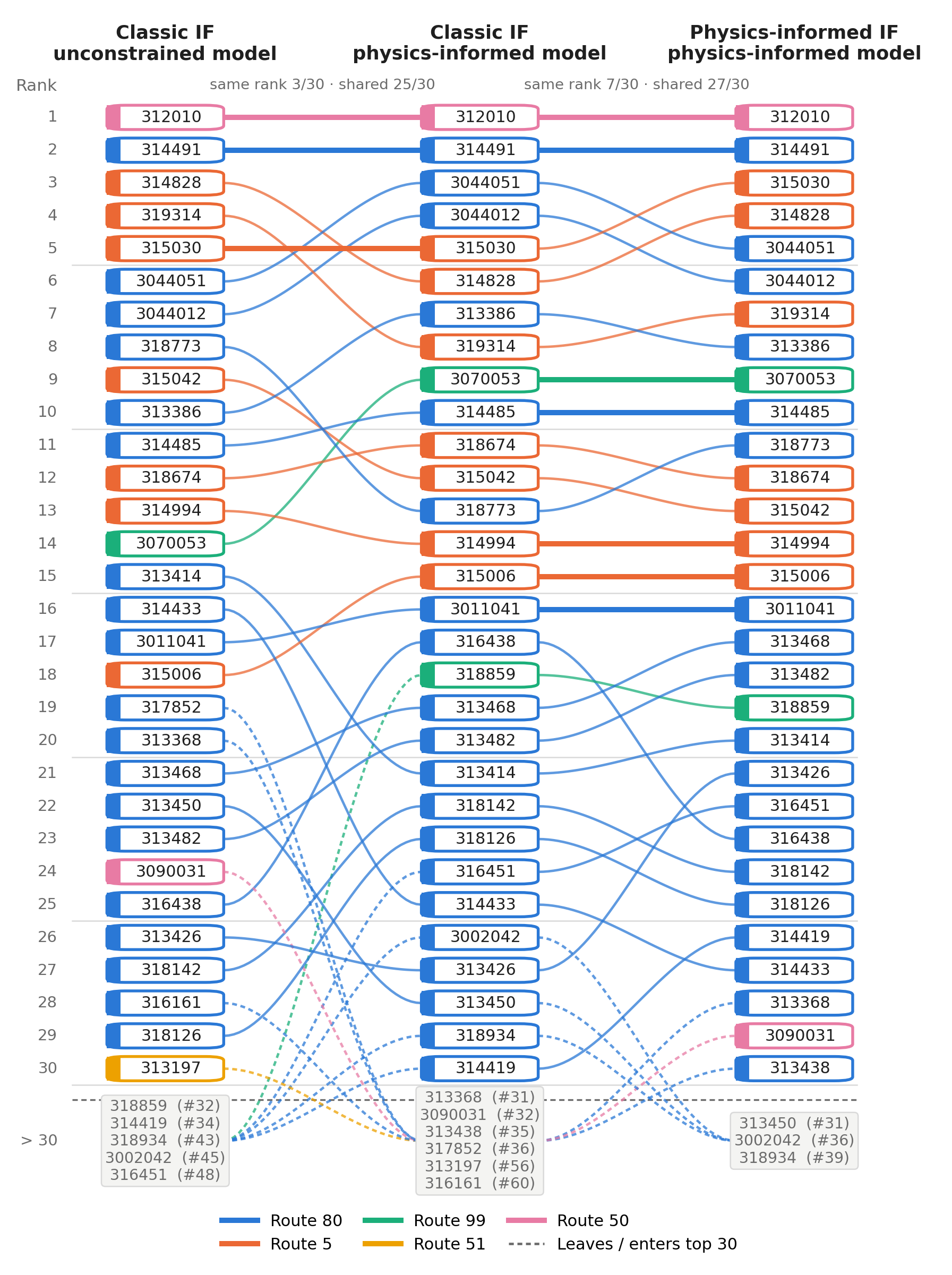}
    \caption{Top 30 stations by importance score under three frameworks:
    classic influence function on the unconstrained model (left), classic influence
    function on the physics-informed model (middle), and physics-informed influence
    function on the physics-informed model (right). Each box is a station at its rank,
    and each line follows the same station from one framework to the next. Colors
    show the freeway route. Thick flat lines mark stations with the same rank in both
    frameworks. Dashed lines mark stations that enter or leave the top 30; their rank
    outside the top 30 is shown at the bottom.}
    \label{fig:top30_rank}
\end{figure}

Figure~\ref{fig:top30_rank} compares the top 30 most important stations based on the SSS score calculated under the three frameworks. Each column corresponds lists the top 30 stations in the order of importance under one framework.  The lines flow between the columns show the varying order of importance across different frameworks. The top two stations, 312010 on Route 50 and 314491 on Route 80, are consistently identified as the most important across all three methods. However, incorporating physics constraints into the model and influence analysis leads to different influence estimates, substantially altering the relative rankings of most sensors in the top 30.

\subsection{Comparison with Traffic-based and Topological Importance Measures}

We now examine whether the physics-informed influence score (SSS) provide new information beyond simpler importance measures such as traffic volume, node degree centrality, and betweenness centrality.  The node degree is the number of road links that converge at a station. Betweenness centrality measures how frequently the station lies on shortest paths of demand; the higher the value, the more central the station is.

\begin{figure}[htbp]
\centering
\begin{subfigure}[b]{0.49\textwidth}
  \centering
  \includegraphics[width=\textwidth,  height=0.28\textheight]{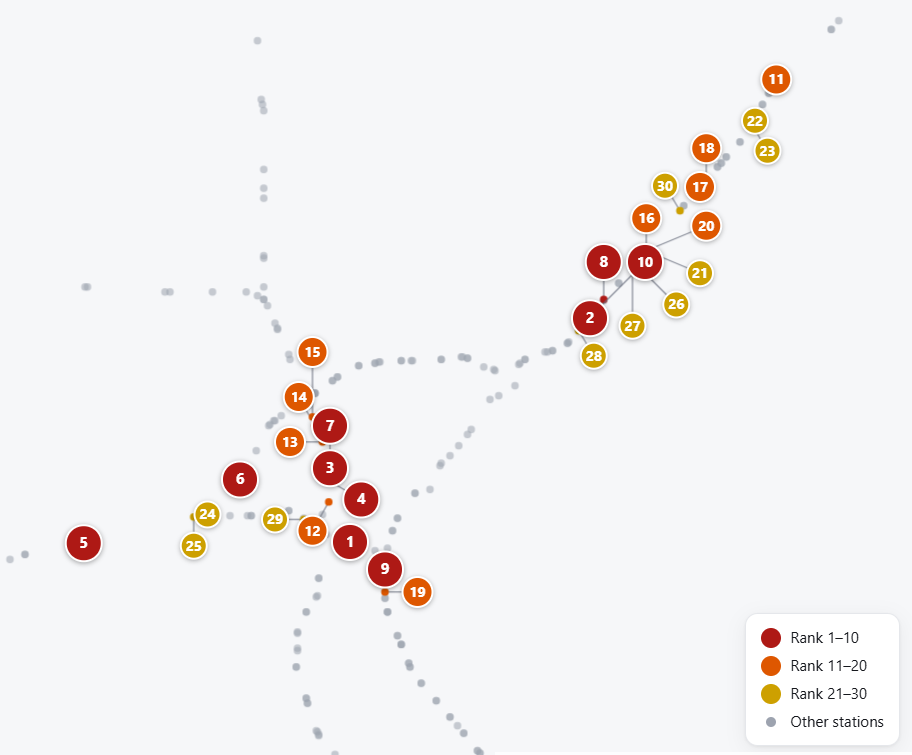}
  \caption{SSS (physics-informed influence)}
  \label{fig:top30_sss}
\end{subfigure}
\hfill
\begin{subfigure}[b]{0.49\textwidth}
  \centering
  \includegraphics[width=\textwidth, height=0.28\textheight]{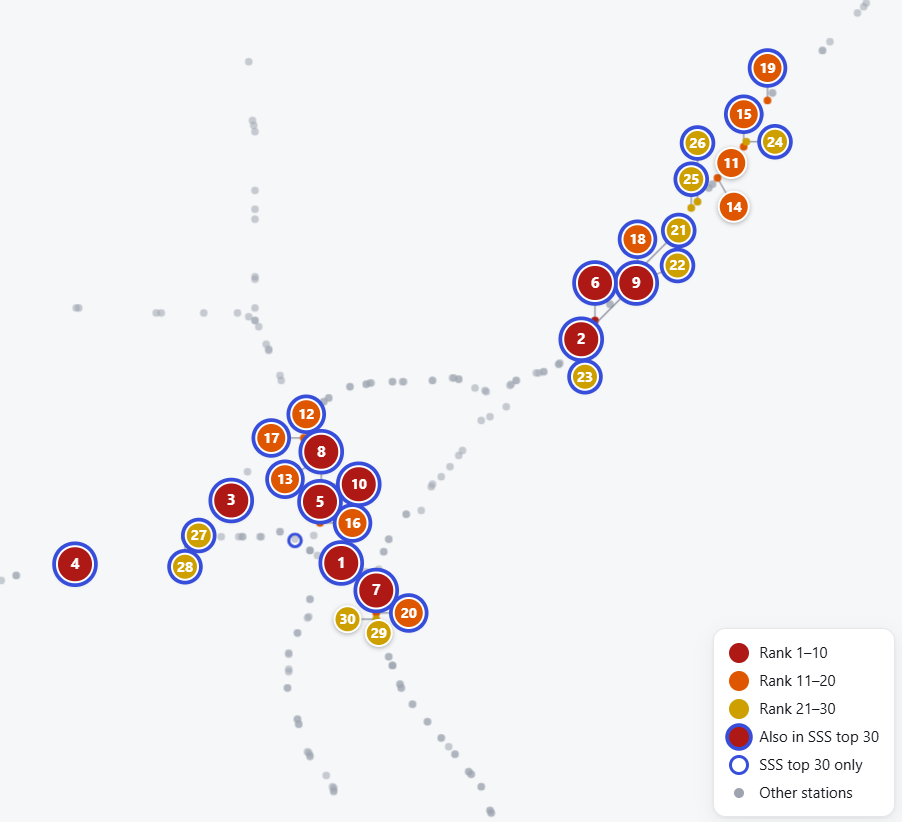}
  \caption{Traffic volume (26 of 30 shared with SSS)}
  \label{fig:top30_flow}
\end{subfigure}

\vspace{0.4em}
\begin{subfigure}[b]{0.49\textwidth}
  \centering
  \includegraphics[width=\textwidth, height=0.28\textheight]{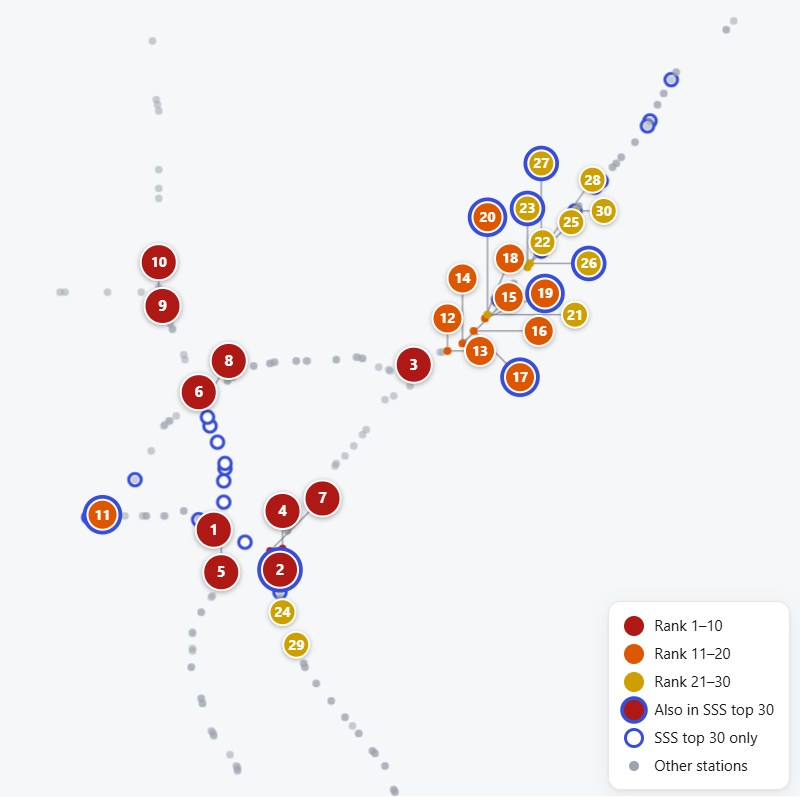}
  \caption{Degree (8 of 30 shared with SSS)}
  \label{fig:top30_degree}
\end{subfigure}
\hfill
\begin{subfigure}[b]{0.49\textwidth}
  \centering
  \includegraphics[width=\textwidth, height=0.28\textheight]{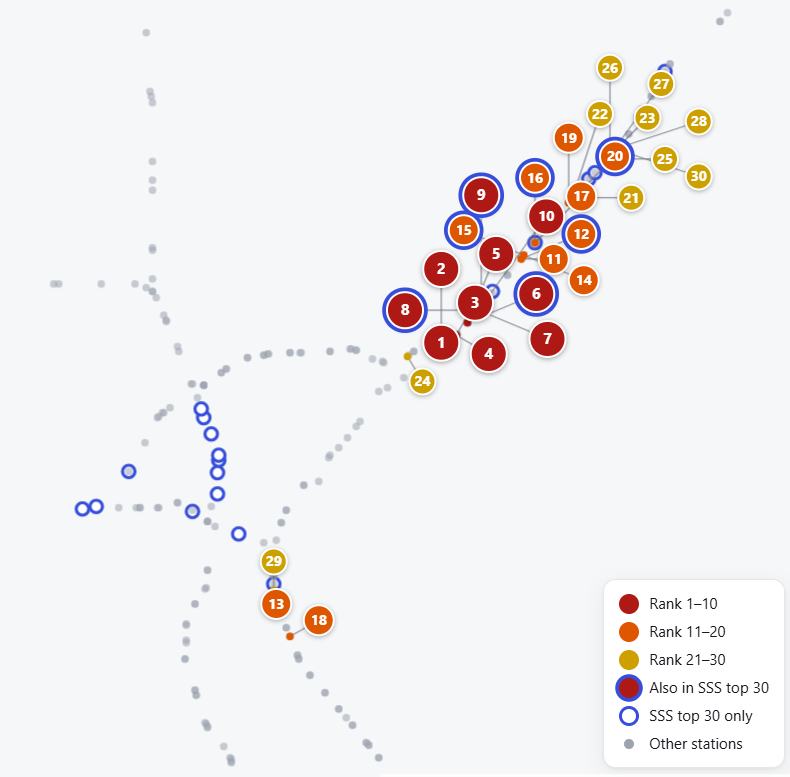}
  \caption{Betweenness (7 of 30 shared with SSS)}
  \label{fig:top30_btw}
\end{subfigure}
\caption{Top 30 stations by (a) the physics-informed influence score (SSS),
(b) mean hourly traffic volume, (c) degree, and (d) betweenness centrality.
Red markers show the top 30 stations of each measure; blue rings mark stations
that are also in the SSS top 30. The table beside each map lists the ranking
and the SSS rank of each station.}
\label{fig:top30_benchmarks}
\end{figure}

Figure \ref{fig:top30_benchmarks} shows the top 30 stations under each measure. The two network-centrality measures point to very different stations compared with SSS. Betweenness selects a continuous stretch, where many shortest paths pass, but only 7 of these stations are in the SSS top 30. Several of them rank below 200 by SSS. Degree performs no better, with 8 shared stations. The 11 stations with degree 3 or 4 are all at freeway-to-freeway interchanges. Only two of them are in the SSS top 30, and the rest rank between 83 and 290. Degree cannot separate the remaining stations, since most of the stations have exactly two road links. Traffic volume is closer to SSS; 26 out of the 30 stations appear in the top 30 list from both methods. This is expected, because busier stations contribute more to network-wide prediction error. However, the ranking of the 40 sensors differ substantially. The two lists also differ at the top-ranked stations. Stations 318934 and 3002042 carry high volume (ranks 11 and 14), but SSS places them outside its top 30 (ranks 39 and 36). The differences suggest that physics-informed influence provides information that is not readily captured by conventional traffic-based or topological metrics. 

\subsection{Assessing the Learning Impact of Influential Sensors}
\label{sec:holdout}

To evaluate whether SSS accurately identifies stations that are important for model learning, we conduct a sensor holdout experiment using the unconstrained baseline model and HardNetAff-STGNN model. Two separate experiments were performed: withholding data from
the ten highest-ranked stations and withholding data from the ten
lowest-ranked stations during training. Each model was retrained
under both conditions and evaluated across all 318 stations using MAE, RMSE, and $R^2$. The Percentage changes were calculated relative to the corresponding full model trained using all stations. 

In each trial, the input features for the removed stations are masked in the training set. The model is retrained from scratch and tested on all 318 stations using the full unmasked test set. 

As shown in table~\ref{tab:holdout_comparison},Withholding the top-10 SSS stations substantially degrades both models. For the
unconstrained model, MAE and RMSE increase by approximately
52.6\% and 119\%, respectively, while $R^2$ decreases by 33.2\%.
The physics-informed model shows increases of 44.2\% in MAE and 114.19\% in RMSE, together with a 32\% reduction in $R^2$.
In contrast, withholding the bottom-10 stations increases MAE
and RMSE by approximately 4\% and 2.1\% for the
unconstrained model, while reducing these errors by 1.87\%
and 0.76\% for the physics-informed model. The corresponding
Changes in $R^2$ are near zero for both models.
These results explain the usefulness of SSS for identifying stations whose data strongly influence the prediction accuracy. Although the physics-informed model shows smaller relative increases in error following top-10 removal, both models remain sensitive to the loss of these stations.

\begin{table*}[ht]
\centering
\caption{Impact of holding out the top-10 and bottom-10 SSS stations
on the unconstrained and physics-informed models.
All metrics are evaluated across all stations.
$\Delta$ (\%) denotes the percentage change relative to
the corresponding full model.}
\label{tab:holdout_comparison}
\resizebox{\textwidth}{!}{%
\begin{tabular}{lcccccccc}
\toprule
& \multicolumn{4}{c}{Unconstrained}
& \multicolumn{4}{c}{Physics-informed} \\
\cmidrule(lr){2-5}\cmidrule(lr){6-9}
& \multicolumn{2}{c}{Holdout value}
& \multicolumn{2}{c}{$\Delta$ (\%)}
& \multicolumn{2}{c}{Holdout value}
& \multicolumn{2}{c}{$\Delta$ (\%)} \\
\cmidrule(lr){2-3}\cmidrule(lr){4-5}
\cmidrule(lr){6-7}\cmidrule(lr){8-9}
Metric
& Top-10 & Bottom-10
& Top-10 & Bottom-10
& Top-10 & Bottom-10
& Top-10 & Bottom-10 \\
\midrule
MAE (veh/hr)
& 383.99 & 263.84 & $+52.6$ & $+4$
& 389.83 & 265.23 & $+44.2$ & $-1.87$ \\
RMSE (veh/hr)
& 945.92 & 439.26 & $+119$ & $+2.1$
& 947.53 & 438.90 & $+114.19$ & $-0.76$ \\
$R^2$
& 0.613 & 0.917 & $-33.2$ & $\approx 0$
& 0.6113 & 0.9166 & $-32$ & $\approx 0$ \\
\bottomrule
\end{tabular}%
}
\end{table*}

The differences between the two removal experiments confirm that SSS is a valuable sensitivity indicator. The ten highest-SSS stations, focused on the downtown Sacramento interchange and along the inbound I-80 corridor, greatly enhance the model's ability to learn flow patterns across the network. Removing them cuts off the model's access to the most limited point in the network, making predictions in downstream areas unreliable. In contrast, the ten lowest-SSS stations are located at the edges and have little impact on the gradient directions learned by the model, so their removal hardly affects its quality.


\section{Conclusions and Discussion}
This paper presents a new framework for evaluating influence of data on transportation network problems subject to linear set of physical constraints like flow conservation.  Through sensitivity analysis and orthogonal decomposition, we provide analytical insights how physical constraints alter the magnitude, direction, and spatial distribution of training-sample influence.  Numerically, we demonstrate the difference between our influence result compared with classical methods, and how such difference translates to model prediction performance.  Practically, Station Sensitivity Score (SSS) based on the influence calculation of each station data can further support downstream applications
to data diagnostics, sensor evaluation, and targeted data collection.

For future research, despite the computational improvements achieved by the proposed framework, extending influence analysis to larger transportation networks remains challenging, particularly as the number of model parameters, monitoring stations, and physical constraints increases. An important direction for future research is to exploit the sparse and localized structure of traffic network interactions to further reduce the computational cost of influence estimation, rather than treating all model parameters and physical constraints as globally coupled.  From a modeling perspective, future studies can relax the assumption of a fixed conservation structure by introducing time-changing constraints that respond to incidents, control actions, or changing demand patterns.   More broadly, the proposed framework lays the groundwork for adaptive, physics-aware learning in transportation networks. By incorporating SSS into online sensing and model updating, future systems could continuously identify the most informative data sources and adapt sensing strategies in response to changing traffic conditions. This perspective may extend the proposed framework beyond static prediction toward closed-loop, self-improving transportation intelligence.

 \ACKNOWLEDGMENT{This work is supported by the National Science Foundation under grant CMMI-2432336/CMMI-2432337.}

\bibliographystyle{plainnat}   
\bibliography{references}   

\end{document}